\documentclass{article}

    \PassOptionsToPackage{numbers, compress}{natbib}
\usepackage[preprint]{neurips_2024}

\usepackage[utf8]{inputenc} % allow utf-8 input
\usepackage[T1]{fontenc}    % use 8-bit T1 fonts
\usepackage{hyperref}       % hyperlinks
\usepackage{url}            % simple URL typesetting
\usepackage{booktabs}       % professional-quality tables
\usepackage{amsfonts}       % blackboard math symbols
\usepackage{nicefrac}       % compact symbols for 1/2, etc.
\usepackage{microtype}      % microtypography
\usepackage{xcolor}         % colors

\usepackage{amssymb}
\usepackage{bm}
\usepackage{amsmath}
\usepackage[ruled,linesnumbered]{algorithm2e}
\usepackage{color}
\usepackage{array}
\usepackage{multirow}
\usepackage{ragged2e}
\usepackage{tabularray}
\usepackage{tikz}
\usepackage{pgfplots}
\usepackage{ulem} 
\usepackage{graphicx}
\usepackage{subfig}
\usepackage{float}
\usepackage{graphbox}

\title{CLIP model is an Efficient Online Lifelong Learner}

\author{%
  Leyuan Wang \\
  Beijing University of \\Posts and Telecommunications \\
  \texttt{leyuan.wang@bupt.edu.cn} \\
  \And
  Liuyu Xiang\thanks{Corresponding author.} \\
  Beijing University of \\Posts and Telecommunications \\
  \texttt{xiangly@bupt.edu.cn} \\
  \And
  Yujie Wei \\
  Wuhan University \\
  \texttt{wyj@whu.edu.cn} \\
  \And
  Yunlong Wang\\
  Institute of Automation,\\ 
  Chinese Academy of Sciences \\
  \texttt{yunlong.wang@cripac.ia.ac.cn} \\
  \And
  Zhaofeng He \\
  Beijing University of \\Posts and Telecommunications \\
  \texttt{zhaofenghe@bupt.edu.cn} \\
}

\begin{document}

\maketitle

\begin{abstract}
Online Lifelong Learning (OLL) addresses the challenge of learning from continuous and non-stationary data streams. Existing online lifelong learning methods based on image classification models often require preset conditions such as the total number of classes or maximum memory capacity, which hinders the realization of real never-ending learning and renders them impractical for real-world scenarios. In this work, we propose that vision-language models, such as Contrastive Language-Image Pretraining (CLIP), are more suitable candidates for online lifelong learning. We discover that maintaining symmetry between image and text is crucial during Parameter-Efficient Tuning (PET) for CLIP model in online lifelong learning. To this end, we introduce the Symmetric Image-Text (SIT) tuning strategy. We conduct extensive experiments on multiple lifelong learning benchmark datasets and elucidate the effectiveness of SIT through gradient analysis. Additionally, we assess the impact of lifelong learning on generalizability of CLIP and found that tuning the image encoder is beneficial for lifelong learning, while tuning the text encoder aids in zero-shot learning.
\end{abstract}

% 主要参考文献
% Online Class Incremental Learning on Stochastic Blurry Task Boundary via Mask and Visual Prompt Tuning
% online、Stochastic Blurry Task Boundary的setting
% CLIP model is an Efficient Continual Learner
% 如何用CLIP做持续学习

\section{Introduction}
\label{sec:intro}

The deep neural networks in practical applications typically follows a supervised training paradigm on pre-collected datasets. This paradigm has proven remarkably effective in closed or constrained environments where the data distribution remains relatively stable. However, in other scenarios, models are expected to learn from data streams and its distribution evolves over time, as depicted in Figure \ref{subfig:searchindex}, so the traditional training paradigm faces significant challenges. Furthermore, due to constraints such as storage or privacy, it is impractical to retain all data. The direct training of models in such an incremental stream of data can lead to drastic performance degradation. This phenomenon is known as catastrophic forgetting \cite{mccloskey1989catastrophic,GoodfellowMDCB13}. 

To mitigate the challenges posed by non-stationary data distributions, a multitude of continual learning methods have been proposed, aiming to strike a balance between model stability and plasticity. Early continual learning approaches involve segmenting continuous data streams into distinct tasks and employing task identifiers to select task-specific components for classification, which is known as task-incremental learning (TIL) \cite{hsu2018re}. Subsequent developments led to class-incremental learning (CIL) \cite{rebuffi2017icarl, kirkpatrick2017overcoming, DouillardRCC22, wang2022dualprompt}, which addresses the scenario where task identifiers are unavailable, necessitating inference based on class information alone. Futhermore, task-agnostic continual learning (TACL) \cite{BangKY0C21, rabhuTD20} operates without explicit task boundaries, enabling models to be trained online with anytime inference. Despite these advancements, traditional image classification models are often designed for closed-set scenarios and encounter various limitations in lifelong learning. These models typically require continuous modifications to their architecture to accommodate new data \cite{DouillardRCC22}, face mismatches between image features and prototypes \cite{ZhaoXGZX20}, or necessitate the predefinition of a maximum class count \cite{seo2024learning}. Such constraints impede their ability to adapt to the ever-evolving nature of real-world data, where new classes and examples are continuously introduced without prior knowledge of their existence. The inability to dynamically evolve with new information challenges the notion that these models are capable of never-ending learning or lifelong learning, thereby restricting their applicability in practical scenarios where the learning process must be sustained and uninterrupted. Consequently, there is a pressing need for models that can transcend these limitations and provide a more flexible, scalable, and realistic approach to learning in open and dynamic environments.

% 可能需要一张类似于lifelong learning的图，介绍分类模型与CLIP匹配的不同

In this paper, we explore the potential of the vision-language models such as Contrastive Language-Image Pretraining (CLIP) model in the scenario of Online Lifelong Learning (OLL), offering a unique approach to classification that transcends traditional model architecture and class count limitations. Unlike conventional classifiers, the CLIP model achieves categorization by matching images to textual descriptions in the form of "\textit{A bad photo of {CLASS}}". This mechanism allows a flexible learning process that is not confined by a predefined model structure or the total number of classes, thereby facilitating realistic OLL. The pre-trained CLIP model, with its generalizable representations and robust zero-shot performance, presents a significant advantage for OLL scenarios. 

To enhance the performance of CLIP in OLL scenarios, we adopt Parameter-Efficient Tuning (PET), an approach that optimizes the model without a significant increase in parameters. Our experiments reveal that the asymmetry between image and text during tuning, where text from all seen classes is matched with images from the current time step, can lead to severe catastrophic forgetting. To address this issue, we introduce the Symmetric Image-Text (SIT) tuning strategy, a straightforward yet potent method that ensures a balanced update of the knowledge of CLIP model. We conduct comprehensive experiments on various lifelong learning benchmark datasets and elucidate the effectiveness of SIT through gradient analysis, providing insights into how symmetric tuning mitigates the loss of previously acquired knowledge. Additionally, we assess the impact of lifelong learning on  generalizability of CLIP model and discover that tuning the image encoder is conducive to incremental learning in OLL, while tuning the text encoder enhances zero-shot learning capabilities. These findings underscore the importance of SIT in maintaining the balance between adapting to new information and retaining existing knowledge, thereby optimizing the performance of CLIP model in the complex and ever-changing landscape of OLL.

\section{Related Work}
% 分两部分，首先从CIL开始，过度到Online CIL，然后讲si-blurry.
    % CIL的部分中，基于动态结构的方法需要展开点，主要是DER和DyTox这俩能增加分类器的：introduction里面强调了一般CIL需要预知总类别数算不上never-dending。
    % 标注的引用需要在dblp上查一下bibtex然后引用上，CIL的三类经典方法可以先跳过
% \wyj{Need to use word "CIL" instead of "CL"?}
%   还是CL吧，通用一些

\textbf{Lifelong Learning.} To alleviate catastrophic forgetting, various lifelong learning methods have been proposed. Existing lifelong methods can be categorized into three types  \cite{abs-2302-03648}: model-based, data-based \cite{rebuffi2017icarl, BangKY0C21}, and algorithm-based \cite{kirkpatrick2017overcoming,li2017learning,hou2019learning,zhu2021prototype} methods. Model-based methods dynamically expand the network for different tasks. \citet{DouillardRCC22} assigns specialized decoder networks and classifiers to each task. \citet{yan2021dynamically} adds a new learnable module composed of a feature extractor and a linear classifier to accommodate new classes at each step. Recently, pioneering works \cite{wang2022learning,wang2022dualprompt} utilize pre-trained models and introduce prompt-tuning to balance the stability and plasticity of the model, demonstrating the superiority of pre-trained models in CL. However, most CL research requires a predefined maximum number of classes to determine the output dimension of the classifier and mainly focuses on the offline scenario, which has clear task boundaries, unlike in the real world. Therefore, online learning with blurry task boundaries \cite{BangKY0C21,rabhuTD20} is proposed, where clear task boundaries are often absent and learned samples are unavailable. Though later \cite{KohK0C22iblurry} proposes a more realistic setting to further tackle the problem of class emerging, recurring and disappearing in the real world, each of its tasks has the same number of new classes and the same ratio between new and blurry classes. To address this, \citet{MoonPKP23siblurry} randomly assigns each blurry class and disjoint class to each blurry task and disjoint task by the disjoint class ratio and every training task is composed of blurry task and disjoint task with a stochastic blurry task boundary, better aligning with the needs of real-world scenarios.

% 第二部分讲CLIP和PET，CLIP的主要包括coop、cocoop和maple，其他PET大概讲讲adapter和LoRA
\textbf{Parameter-Efficient Tuning in CLIP.} Trained with abundantly available data from the web, the vision-language model CLIP \cite{RadfordKHRGASAM21} has demonstrate great advantage in a wide variety of tasks including few-shot and zero-shot visual recognition. However, how to efficiently adapt them to downstream tasks still remains a challenge. To solve this problem, several parameter-efficient tuning methods have been proposed, roughly categorized into LoRA \cite{HuSWALWWC22}, Adapter \cite{houlsby2019parameter}  prompt-tuning \cite{LesterAC21} and prefix-tuning \cite{LiL20} . Inspired by prompt learning in NLP, many works tune CLIP through the learnable prompt. \citet{ZhouYLL22} applies prompt learning-based approach to CLIP for the first time and shows exceptional performance in downstream transfer learning. \citet{ZhouYL022} extends \citet{ZhouYLL22} by learning a lightweight network that generates an input-conditional token for each image to further solve the generalization issue. \cite{KhattakR0KK23} improve the alignment between two modalities by projecting textual prompt into visual prompt and embedding them into corresponding encoders. Adapter approaches insert small modules, called adapters, in the pre-trained model. \citet{GaoGZMFZLQ24} introduces CLIP-Adapter, adding a fully connected layer adapter and then merging the additional knowledge with zero-shot prediction based on residual connections. \citet{abs-2211-09623} and \citet{abs-2301-07868} share adapter weights between visual and textual modalities to enhance the cross-modal interaction. LoRA approaches, on the other hand, inject trainable low-rank matrices into the pre-trained model to realize the tuning of the model. \citet{abs-2310-13683} employs LoRA in CLIP model for language adaptation for the first time, considerably reducing the number of trainable parameters.

\section{Methodology}

\begin{figure*}[!ht]
\centering
    \subfloat[US Extreme Weather Search Interest.]{\includegraphics[align=c,width=.45\linewidth]{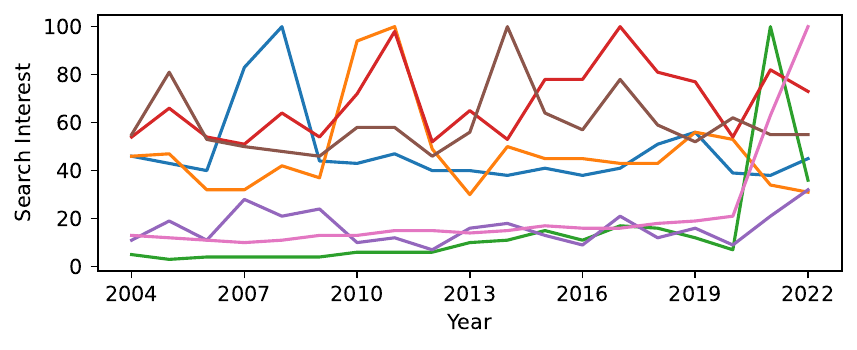}\label{subfig:searchindex}}
    \hspace{5pt}
    \subfloat[Visualization of CIL scenario.]{\includegraphics[align=c,width=.45\linewidth]{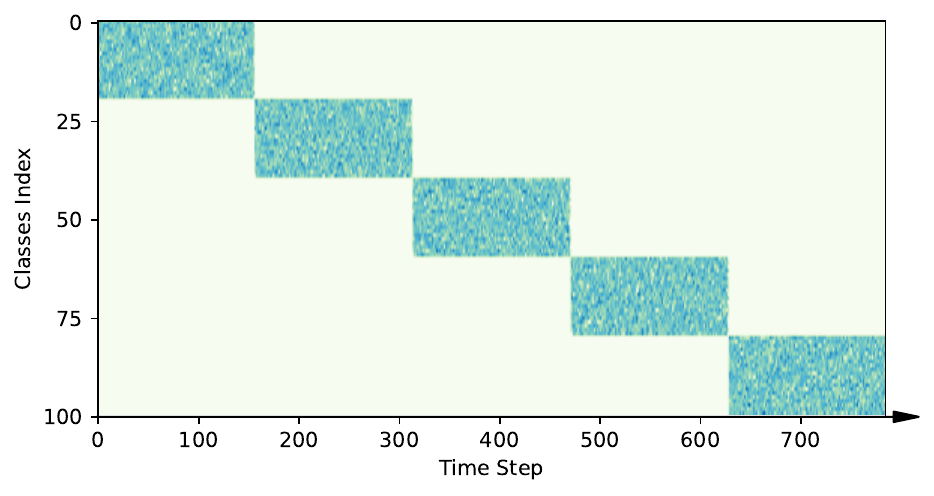}\label{subfig:CIL}}
    \\
    \subfloat[Visualization of i-Blurry scenario.]{\includegraphics[align=c,width=.45\linewidth]{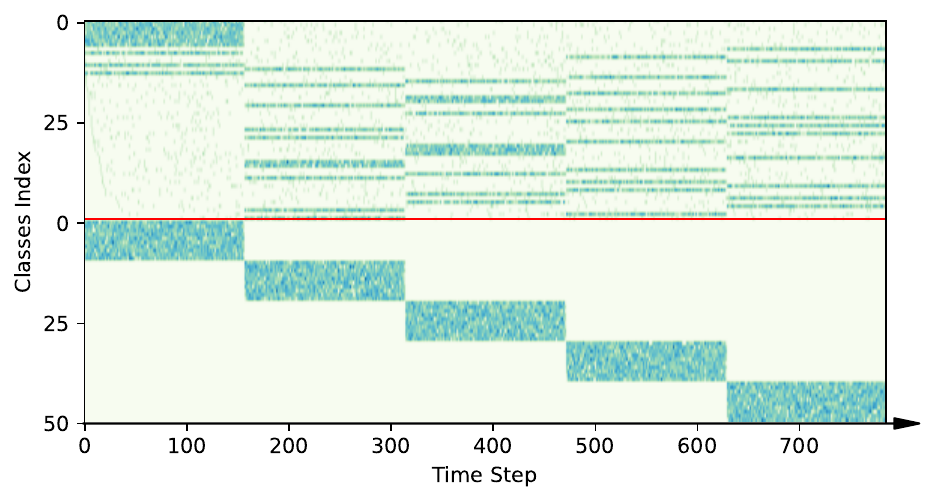}\label{subfig:i_Blurry}}
    \hspace{5pt}
    \subfloat[Visualization of Si-Blurry scenario.]{\includegraphics[align=c,width=.45\linewidth]{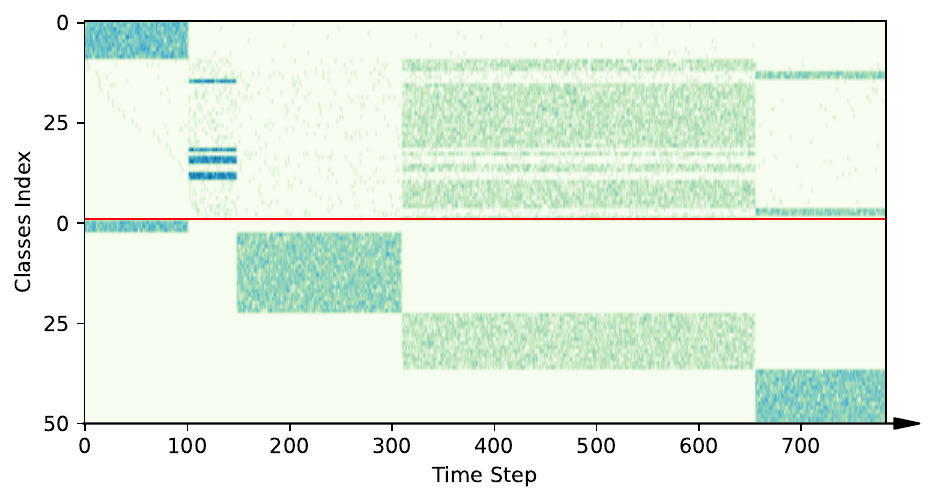}\label{subfig:Si_Blurry}}
    \caption{Data in (a) are from GoogleTrends
    \cite{SearchInterest}. In (b-d), the horizontal coordinate is the time step/batch and the vertical coordinate is the class index. Those above the red solid line are blurry classes, and those below are disjoint classes, both are sorted according to their occurrence time.}
\label{fig:datasteram}
% \vspace{-10pt}
\end{figure*}

\subsection{Problem Formulation}

% 首先参考Continual-CLIP的3.1第一段写一下常规CIL的setting，然后参考i-blurry 第三节第一个黑体部分写一下i-Blurry-N-M Split，最后参考si-blurry的3.1写一下Stochastic的setting
% 需要强调The goal of Online Lifelong Learning is to train a unified model on data seen only once while predicting well on seen classes.

Lifelong learning aims to train a unified model $\mathcal{F}_\theta:\mathcal{X}\rightarrow\mathcal{Y}$ parameterized by $\theta$ that makes good predictions for all seen classes.
% , while only being allowed to see the training data once. 
In classic class-incremental learning setting, given a sequence of tasks $\mathcal{T}=\{\mathcal{T}_1,\mathcal{T}_2,...,\mathcal{T}_T\}$, the training set of the $t^{th}$ task $\mathcal{T}_t$ is $\mathcal{D}_t={(\bm{x}^t_i,y_i^t)}^N_{i=1}$, where $\bm{x}^t_i\in\mathcal{X}$ and $y_i^t\in\mathcal{Y}$ denote the input sample and its corresponding label respectively. We define the output space for all observed class labels $\mathcal{Y}^{(t)}\subset\mathcal{Y}^{(t+1)}$. The label $\mathcal{Y}^{(t)}$ between any two tasks are disjoint, i.e. $\mathcal{Y}^{(t)} \cap \mathcal{Y}^{(t^{'})} = \emptyset$. 
% Typically, in CIL, we have $P(\mathcal{X}^{(t)},\mathcal{Y}^{(t)}) = P(\mathcal{X}^{(t^{'})},\mathcal{Y}^{(t^{'})}))$, where $P(\cdot)$ represents the distribution of the data.

In online lifelong learning, the training data is partitioned into batches $\mathcal{B}=\{\mathcal{B}_1,\mathcal{B}_2,...\}$, the training set of the $t^{th}$ batch $\mathcal{B}_t$ is $\mathcal{D}_t={(\bm{x}^t_i,y_i^t)}^N_{i=1}$. Each batch $\mathcal{B}_i$ is allowed to be seen only once. Here we still use task $\mathcal{T}$ to represent abrupt changes in data distribution, but the learner is unaware of the task alteration during training. For Si-Blurry setting \cite{MoonPKP23siblurry}, we split all classes into two groups, where $N$\% of the classes are selected as disjoint classes and the remaining 100-$N$\% of the classes are selected as blurry$M$ classes, where $M$ is the blurry level \cite{KohK0C22iblurry}. We randomly assign a certain number of disjoint and blurry classes to each task, as shown in Figure \ref{subfig:Si_Blurry}, the blurry classes of each task may overlap, thus blurring the clear task boundaries. In particular, when $N = 0$, all classes are disjoint classes, representing the classical CIL setting, as illustrated in Figure \ref{subfig:CIL}. Conversely, when the ratio of disjoint classes to blurry classes remains constant in each task, it corresponds to the i-Blurry setting \cite{KohK0C22iblurry}, as depicted in Figure \ref{fig:datasteram}.

\subsection{CLIP model as a lifelong learner}
% 参考Continual-CLIP，但需要补充训练text范围

The CLIP model represents a significant advancement in multimodal machine learning. The CLIP model is designed to map images and text into a shared feature space, with the pre-training objective of maximizing the similarity between feature vectors of image-text pairs. This approach contrasts with traditional methods that focus on minimizing the similarity between non-matching pairs. 
% The CLIP model lies in its ability to learn rich joint representations, and it can captures the nuanced relationships between visual content and textual descriptions, which is pivotal for tasks that involve understanding and generating language conditioned on images. 
Consider a $K$-class image classiﬁcation problem, CLIP maps an unidentified image $\bm{x} \in \mathcal{X}$ to its corresponding feature vector through the image encoder $\bm{v}=\mathbf{E}_\texttt{visual}(\mathbf{x})$. Class label $y \in \mathcal{Y}$ is prepended by a hand-crafted prompt template $\mathbf{p} \rightarrow$ \textit{a bad photo of a \{class\}} to form a class-specific text input $\mathbf{y} = \{\mathbf{p};y\}$, which is then encoded into a text feature vector $\bm{t}$ by the text encoder $\bm{t}=\mathbf{E}_\texttt{text}(\mathbf{y})$. The prediction probability can be denoted by 
\begin{equation}
p(y_{i} |\bm{x}) = \frac{\text{exp}(\text{sim}(\bm{t}_i\cdot\bm{v}))}{\sum_{i=1}^{K}\text{exp}(\text{sim}(\bm{t}_i\cdot\bm{v}))}, 
\end{equation}
\noindent where $\text{sim}(\cdot)$ denotes the cosine similarity.

Traditional image classification models are constrained by a predefined set of classes, which necessitates model retraining or adjustments when novel classes are introduced in lifelong learning. In contrast, the design of CLIP model overcomes these limitations by employing a text-based approach. This approach allows CLIP to dynamically adapt to new classes without altering the model architecture. The incorporation of textual descriptions as classifiers provides a flexible and scalable solution for lifelong learning. This capability is particularly advantageous in environments where the set of classes is continuously expanding.

\subsection{Symmetric Image-Text tuning strategy}
\label{sect:sit}

The vast dataset provides CLIP with a broad understanding of visual and textual patterns, enabling it to generalize well across a wide range of classes. However, when facing the case where the data is not covered by the pre-trained dataset, the performance can falter. Therefore, it is imperative to develop PET strategies for the CLIP model that enhance its performance on specialized tasks while maintaining its robust zero-shot learning abilities, ensuring that the model remains versatile and adaptable in a lifelong learning environment. After adding adapter or prefix to the CLIP model, it is typical to tune with the InfoNCE loss
\begin{equation}
\mathcal{L} = - \sum_{\substack{\bm{v_i} \in V_b}} \log \frac{\text{exp}(\text{sim}(\bm{v_i},\bm{t^{+}}) / \tau)}{\sum_{\bm{t_j} \in T} \text{exp}(\text{sim}(\bm{v_i},\bm{t_j}) / \tau)},
\end{equation}
\noindent where $\bm{v_i}$ and $\bm{t^{+}}$ are the positive sample pairs, $V_b$ is the image feature vectors of the current batch, and $T$ is the text feature vectors of all seen classes.
Unlike the pre-training phase of CLIP model, online lifelong learning scenarios limit the access to the current batch of images only, with text features drawn from all previously seen classes. Our experiments  in \ref{subsec:sit} have revealed that during tuning, the asymmetry between image and text causes the gradient of the negative samples to be smaller than the gradient of the positive samples, causing the model to tend to predict the old class as the new class, resulting in a significant performance degradation. To counteract this issue, we propose a straightforward yet effective method, called Symmetric Image-Text (SIT) tuning strategy, that restores the symmetry between image and text features during training. Specifically, we reformulate the loss function
\begin{equation}
\mathcal{L} = - \sum_{\substack{\bm{v_i} \in V_b}} \log \frac{\text{exp}(\text{sim}(\bm{v_i},\bm{t^{+}}) / \tau)}{\sum_{\bm{t_j} \in T_{b}} \text{exp}(\text{sim}(\bm{v_i},\bm{t_j}) / \tau)},
\end{equation}
\noindent where $T_b$ is the text feature vectors of classes in the current batch.
By doing so, we effectively mitigate catastrophic forgetting, allowing the model to maintain its zero-shot learning capabilities while adapting to new classes in an online lifelong learning context.

\section{Experiments and Results}

\subsection{Experimental Details}

\textbf{Datasets.} We conducted comprehensive experiments on several benchmark datasets to evaluate the performance of our strategy. For OLL and CIL, we selected datasets such as CIFAR-100 \cite{krizhevsky2009learning}, Tiny-ImageNet \cite{le2015tiny}, and ImageNet-R \cite{hendrycks2021many}, which consist of 100, 200, and 200 classes. In the OOL scenario, we employ the Si-blurry setting, and set a disjoint class ratio $N=50$, a blurry level $M=10$ and a task number $T=5$. To assess the impact of OLL on the generalization of CLIP model, we tested on five distinct datasets. These include the generic-objects dataset Caltech101 \cite{fei2004learning}, fine-grained datasets such as Flowers102 \cite{nilsback2008automated}, OxfordPets \cite{parkhi2012cats}, Food101 \cite{bossard2014food}, and a satellite image dataset EuroSAT \cite{helber2019eurosat}. 

\textbf{Implementation Details.} We conducted experiments on a pre-trained ViT-B/16 CLIP model, where $d_l=512$, $d_v=768$ and $d_{vl}=512$. The prompt template utilizes "\textit{a bad photo of a \{class\}}". For the implementation of LoRA \cite{HuSWALWWC22} and Adapters \cite{houlsby2019parameter}, we referred to \citet{he2021towards}. The Adapters featured a down-projection dimension of 64, while the rank of LoRA is set to 4. Unless specified otherwise, both Adapters and LoRA are integrated into each transformer layer of the Text and Image encoders. We employ the Adam optimizer for tuning the CLIP model with either Adapters or LoRA, using a learning rate of $5e-4$ and training for 3 iterations per batch. For the implementation of MaPLe \cite{KhattakR0KK23}, we set the prompt depth $J$ to 9 and standardized the lengths of both the language and vision prompts to 2. MaPLe is optimized using the SGD optimizer with a learning rate of $1e-4$, also over 3 iterations per batch.

\textbf{Evaluation Metrics.} We record the top-1 accuracy $\mathcal{A}_t$ of the model on the test set after finishing training at incremental step $t$ and present it as a curve, where the test set contains all classes the model has ever seen.  We denote the accuracy at the end of the last task $\mathcal{A}_{\text{last}}$ as a metric for overall accuracy and use the average of the test accuracy across all incremental steps $\mathcal{A}_{\text{avg}} = \frac{1}{T} \sum_{t=0}^{T-1} \mathcal{A}_t$ to assess the performance of all tasks. Furthermore, to evaluate the online learning ability of the model, we also use $\mathcal{A}_{\text{auc}}$ \cite{KohK0C22iblurry} to measure the performance of anytime inference, which assumes that inference queries can be made anytime during training. $\mathcal{A}_{\text{auc}}=\sum^{k}_{i=1}{f_{A}(i \cdot \delta n)} \cdot \delta n$, where $\delta n$ is the number of seen samples during the evaluation and $f_{A}(\cdot)$ is the accuracy curve.

% 1. Si-Blyrry性能比较（cifar100、tinyimagenet，imagener-r，task=5/10，N=50，M=10），与pre-train vit比较+CLIP-zeroshot+几种PET的比较
\begin{table}[!htb]
\centering
\caption{Comparison of state-of-the-art online lifelong learning methods in Si-Blurry setting with disjoint class ratio $N=50$, blurry level $M=10$ and task number $T=5$. Note that the buffer size of the second group of methods is 2000, and the other methods are rehearsal-free.}
\label{tab:si-blurry}
\scalebox{0.85}{
\begin{tblr}{
  width = \linewidth,
  colspec = {Q[270]Q[125]Q[125]Q[125]Q[125]Q[125]Q[125]},
  cells = {c},
  row{1} = {c},
  row{2} = {c},
  column{1} = {r},
  cell{1}{1} = {r=2}{},
  cell{1}{2} = {c=2}{0.268\linewidth},
  cell{1}{4} = {c=2}{0.262\linewidth},
  cell{1}{6} = {c=2}{0.25\linewidth},
  hline{1,18} = {-}{0.1em},
  hline{3} = {-}{},
  hline{5,10,14} = {-}{dashed},
  hline{2} = {2-3}{leftpos = -1, rightpos = -1, endpos},
  hline{2} = {4-5}{leftpos = -1, rightpos = -1, endpos},
  hline{2} = {6-7}{leftpos = -1, rightpos = -1, endpos},
}
\textbf{ Method } & CIFAR-100 &  & TinyImageNet &  & ImageNet-R & \\
 & $\mathcal{A}_{\text{auc}}$ & $\mathcal{A}_{\text{last}}$ & $\mathcal{A}_{\text{auc}}$ & $\mathcal{A}_{\text{last}}$ & $\mathcal{A}_{\text{auc}}$ & $\mathcal{A}_{\text{last}}$\\
Finetuning & 19.71±3.39 & 10.42±4.92 & 15.50±0.74 & 10.42±4.92 & 7.51±3.94 & 2.29±0.85\\
Linear Probing & 49.69±6.09 & 23.07±7.33 & 42.15±2.79 & 21.97±6.43 & 29.24±1.26 & 16.87±3.14\\
ER~\cite{rolnick2019experience} & 69.86±4.08 & 71.81±0.69 & 66.75±1.13 & 55.07±1.28 & 45.74±1.35 & 38.13±0.32\\
EWC++~\cite{kirkpatrick2017overcoming} & 47.75±5.35 & 46.93±1.44 & 64.92±1.21 & 53.04±1.53 & 30.20±1.31 & 21.28±1.88\\
RM~\cite{BangKY0C21} & 53.27±3.00 & 65.51±0.55 & 47.26±1.13 & 44.55±0.37 & 27.88±1.29 & 24.25±0.99\\
CLIB~\cite{KohK0C22iblurry} & 71.53±2.61 & 72.09±0.49 & 65.47±0.76 & 56.87±0.54 & 42.69±1.30 & 35.43±0.38\\
MVP-R~\cite{MoonPKP23siblurry} & 78.65±3.59 & \textbf{84.42±0.44} & \textbf{80.67±0.75} & 74.34±0.32 & 52.47±1.45 & 50.54±2.08\\
LwF~\cite{li2017learning} & 55.51±3.49 & 36.53±10.96 & 49.00±1.52 & 27.47±7.59 & 31.61±1.53 & 20.62±3.67\\
L2P~\cite{wang2022learning} & 57.08±4.43 & 41.63±12.73 & 52.09±1.92 & 35.05±5.73 & 29.65±1.63 & 19.55±4.78\\
DualPrompt~\cite{wang2022dualprompt} & 67.07±4.16 & 56.82±3.49 & 66.09±2.00 & 48.72±3.41 & 40.11±1.27 & 29.24±4.63\\
MVP~\cite{MoonPKP23siblurry} & 68.10±4.91 & 62.59±2.38 & 68.95±1.33 & 52.78±2.08 & 40.60±1.21 & 31.96±3.07\\
Continual-CLIP~\cite{Thengane22cclip} & 69.57±1.05 & 66.26±0.02 & 71.55±0.97 & 65.18±0.03 & 76.63±0.52 & 71.12±0.34\\
\textbf{SIT-MaPLe} & 69.89±2.69 & 67.61±1.01 & 71.17±1.45 & 65.78±1.90 & 78.50±0.74 & 73.48±1.00\\
\textbf{SIT-Adapter} & \textit{81.03±2.00} & 79.15±2.08 & 80.30±1.19 & \textit{74.96±1.44} & \textit{83.28±1.01} & \textit{78.76±0.39}\\
\textbf{SIT-LoRA} & \textbf{81.77±1.80} & \textit{80.80±1.26} & \textit{80.64±1.18} & \textbf{75.08±1.09} & \textbf{84.15±0.38} & \textbf{79.35±0.35}
\end{tblr}}
\end{table}

\subsection{Experimental Result}

\subsubsection{Online Lifelong Learning.}  

In this experiments, we utilized the Si-Blurry setting to evaluate our proposed method on CIFAR-100, TinyImageNet, and ImageNet-R datasets. Specifically, we set the disjoint class ratio $N=50$, blurry level $M=10$ and task number $T=5$. Our method is compared with several state-of-the-art OLL approaches, including replay-based methods ER~\cite{rolnick2019experience}, RM~\cite{BangKY0C21}, and CLIB~\cite{KohK0C22iblurry}, regularization-based method LwF~\cite{li2017learning}, a combination of replay-regularization in EWC++~~\cite{kirkpatrick2017overcoming}, and prompt-based methods L2P~\cite{wang2022learning}, DualPrompt~\cite{wang2022dualprompt}, MVP and MVP-R~\cite{MoonPKP23siblurry}. Additionally, we considered Continual-CLIP, which leverages the zero-shot learning capability of CLIP model, and set finetuning and linear probing as our lower-bound benchmarks. All methods employ a pre-trained Vision Transformer (ViT-B/16) as the backbone model. For replay-based methods ER, RM, CLIB, and MVP-R, the buffer size is consistently set to 2000 to ensure a fair comparison. The results, as depicted in Table \ref{tab:si-blurry}, reveal that CLIB and MVP, designed specifically for boundary-blurred scenarios, perform well on general datasets like CIFAR-100 and TinyImageNet. However, even with replay data, these classical classification models only marginally outperform Continual-CLIP and significantly underperform on the domain-shift designed ImageNet-R. Furthermore, we found that applying the SIT strategy to tune CLIP model via PET can substantially enhance the performance of CLIP model without the need for replay, knowledge distillation, or other auxiliary techniques. Our findings underscore the effectiveness of our proposed method, which not only matches but also surpasses the performance of existing methods, highlighting its potential as a robust solution for lifelong learning in blurred and dynamic environments.

% 2. 常规CIL性能比较（cifar100、tinyimagenet，task=5，10，20），部分数据照抄其他论文结果

\begin{table}
\centering
\caption{Comparison of state-of-the-art CIL methods on CIFAR-100 benchmark.}
\label{tab:cil_cifar}
\scalebox{0.85}{
\begin{tblr}{
  width = \linewidth,
  colspec = {Q[280]Q[179]Q[179]Q[179]Q[179]},
  cells = {c},
  column{1} = {r},
  cell{1}{1} = {r=2}{},
  cell{1}{2} = {c=2}{0.358\linewidth},
  cell{1}{4} = {c=2}{0.358\linewidth},
  hline{1,16} = {-}{0.1em},
  hline{3} = {-}{},
  hline{12} = {-}{dashed},
  hline{2} = {2-3}{leftpos = -1, rightpos = -1, endpos},
  hline{2} = {4-5}{leftpos = -1, rightpos = -1, endpos},
}
\textbf{Method} & \textbf{10 tasks} &  & \textbf{20 tasks} & \\
 & $\mathcal{A}_{\text{avg}}$ & $\mathcal{A}_{\text{last}}$ & $\mathcal{A}_{\text{avg}}$ & $\mathcal{A}_{\text{last}}$\\
iCaRL~\cite{rebuffi2017icarl}& 65.27 & 50.74 & 61.20 & 43.74\\
UCIR~\cite{hou2019learning} & 58.66 & 43.39 & 58.17 & 40.63\\
BiC~\cite{WuCWYLGF19} & 68.80 & 53.54 & 66.48 & 47.02\\
WA~\cite{ZhaoXGZX20} & 69.46 & 53.78 & 67.33 & 47.31\\
PODNet~\cite{douillard2020podnet} & 58.03 & 41.05 & 53.97 & 35.02\\
DER(w/o P)~\cite{yan2021dynamically} & 75.36 & 65.22 & 74.09 & 62.48\\
DER~\cite{yan2021dynamically} & 74.64 & 64.35 & 73.98 & 62.55\\
DyTox~\cite{DouillardRCC22} & 67.33 & 51.68 & 67.30 & 48.45\\
DyTox+~\cite{DouillardRCC22} & 74.10 & 62.34 & 71.62 & 57.43\\
Continual-CLIP~\cite{Thengane22cclip} & 75.17 & 66.72 & 75.95 & 66.72\\
\textbf{SIT-MaPLe} & 74.83±1.56 & 66.58±1.42 & 74.61±0.93 & 67.31±0.35 \\
\textbf{SIT-Adapter} &  \textit{85.55±0.37} & \textit{77.80±0.80} & \textit{85.98±0.33} & \textit{79.29±1.35} \\
\textbf{SIT-LoRA} & \textbf{86.45±0.40} & \textbf{78.67±0.70}  & \textbf{86.88±0.34} & \textbf{80.38±0.57}
\end{tblr}}
\end{table}

\subsubsection{Class-Incremental Learning.} 

We evaluated the performance of our proposed SIT against several state-of-the-art CILmethods, including EWC~\cite{kirkpatrick2017overcoming}, LwF~\cite{li2017learning}, iCaRL~\cite{rebuffi2017icarl}, EEIL~\cite{castro2018end}, UCIR~\cite{hou2019learning}, BiC~\cite{WuCWYLGF19}, WA~\cite{ZhaoXGZX20}, PODNet~\cite{douillard2020podnet}, MUC~\cite{liu2020more}, DER~\cite{yan2021dynamically}, PASS~\cite{zhu2021prototype}, and DyTox~\cite{DouillardRCC22} on CIFAR-100 and TinyImagenet. These CIL methods adhere to the traditional setup where the model has access to the entire dataset of the current task. For Continual-CLIP~\cite{Thengane22cclip} that do not require tuning, we recorded the test results at the end of each task. In contrast, our SIT strategy follows an online CIL setting, where the model is restricted to access the data in the current batch only, simulating a more challenging and realistic incremental learning scenario. The experimental results for the compared CIL methods are sourced from \citet{Thengane22cclip}, providing a reliable benchmark for our evaluation. As illustrated in Table \ref{tab:cil_cifar}, it is evident that our SIT-LoRA method outperforms all other methods in both task settings. Specifically, in the more challenging 20 tasks setting, SIT-LoRA is impressive with an $\mathcal{A}_{\text{avg}}$ of 86.88 and an $\mathcal{A}_{\text{last}}$ of 80.38. Table \ref{tab:cil_tiny} shows that the proposed SIT methods outperform the baseline and other state-of-the-art methods in this comparison. Specifically, SIT-Adapter shows the highest $\mathcal{A}_{\text{avg}}$ across all task increments, with 81.93 for 5 tasks, 81.31 for 10 tasks, and 80.50 for 20 tasks. It also achieves the highest $\mathcal{A}_{\text{last}}$ for the 20 tasks setting with 72.62. The results demonstrate the efficacy of our SIT strategy, showcasing its potential to compete with or even surpass established CIL methods while adhering to a more stringent online learning constraint.

\begin{table}
\centering
\caption{Comparison of state-of-the-art CIL methods on TinyImageNet benchmark.}
\label{tab:cil_tiny}
\scalebox{0.85}{
\begin{tblr}{
  width = \linewidth,
  colspec = {Q[280]Q[131]Q[131]Q[131]Q[131]Q[131]Q[131]},
  cells = {c},
  column{1} = {r},
  cell{1}{1} = {r=2}{},
  cell{1}{2} = {c=2}{0.262\linewidth},
  cell{1}{4} = {c=2}{0.262\linewidth},
  cell{1}{6} = {c=2}{0.262\linewidth},
  hline{1,15} = {-}{0.1em},
  hline{3} = {-}{},
  hline{11} = {-}{dashed},
  hline{2} = {2-3}{leftpos = -1, rightpos = -1, endpos},
  hline{2} = {4-5}{leftpos = -1, rightpos = -1, endpos},
  hline{2} = {6-7}{leftpos = -1, rightpos = -1, endpos},
}
\textbf{Method} & \textbf{5 tasks}  &  & \textbf{10 tasks} &  & \textbf{20 tasks} & \\
 & $\mathcal{A}_{\text{avg}}$ & $\mathcal{A}_{\text{last}}$ & $\mathcal{A}_{\text{avg}}$ & $\mathcal{A}_{\text{last}}$ & $\mathcal{A}_{\text{avg}}$ & $\mathcal{A}_{\text{last}}$\\
EWC~\cite{kirkpatrick2017overcoming} & 19.01 & 6.00 & 15.82 & 3.79 & 12.35 & 4.73\\
LwF~\cite{li2017learning} & 22.31 & 7.34 & 17.34 & 4.73 & 12.48 & 4.26\\
iCaRL~\cite{rebuffi2017icarl} & 34.27 & 23.22 & 30.94 & 20.82 & 27.83 & 20.16\\
EEIL~\cite{castro2018end} & 47.17 & 35.12 & 45.03 & 34.64 & 40.41 & 29.72\\
UCIR~\cite{hou2019learning} & 50.30 & 39.42 & 48.58 & 37.29 & 42.84 & 30.85\\
MUC~\cite{liu2020more} & 32.23 & 19.20 & 26.67 & 15.33 & 21.89 & 10.32\\
PASS~\cite{zhu2021prototype} & 49.54 & 41.64 & 47.19 & 39.27 & 42.01 & 32.93\\
DyTox~\cite{DouillardRCC22} & 55.58 & 47.23 & 52.26 & 42.79 & 46.18 & 36.21\\
Continual-CLIP~\cite{Thengane22cclip} & 70.49 & 66.43 & 70.55 & 66.43 & 70.51 & 66.43\\
\textbf{SIT-MaPLe} & 72.35±0.66 & 65.88±0.95 & 72.35±0.66 & 65.88±0.95 & 72.98±0.96 & 64.69±1.32\\
\textbf{SIT-Adapter} & \textbf{81.93±0.96} & \textbf{75.72±0.45 } & \textit{81.31±0.24} & \textit{74.29±0.05} & \textit{80.50±0.42} & \textbf{72.62±1.23 }\\
\textbf{SIT-LoRA} & \textit{81.15±0.32 } & \textit{75.19±0.55} & \textbf{81.99±0.36 } & \textbf{74.45±0.62} & \textbf{81.35±0.54} & \textit{73.55±0.54}
\end{tblr}}
\end{table}

\subsection{Ablation Study}

% In this section, we delve into the analysis of the impact of online lifelong learning on the generalizability of CLIP, the efficacy of the SIT tuning strategy from a gradient perspective, and the influence of batch size on the performance of online lifelong learning.

\subsubsection{Analysis of Symmetric Image-Text tuning strategy.}
\label{subsec:sit}

\begin{figure*}[!ht]
\centering
    \subfloat[Confusion matrix (AIT).]{\includegraphics[align=c,width=.3\linewidth]{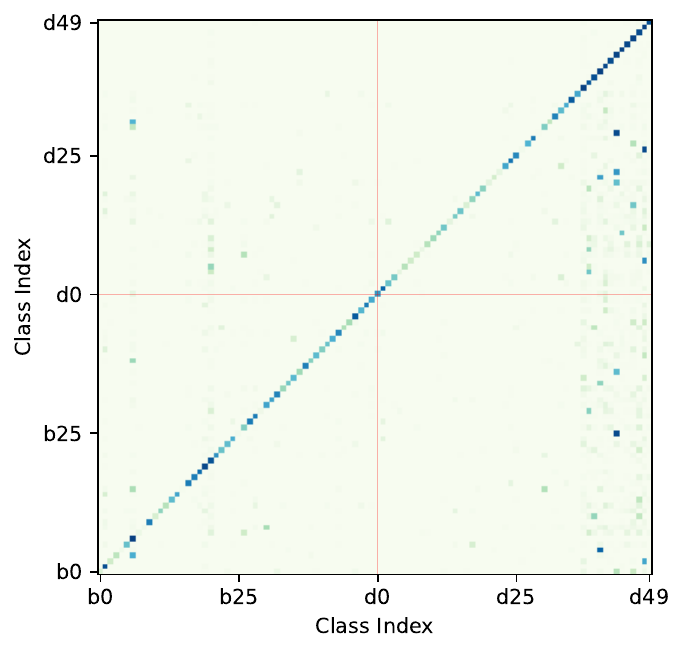}\label{subfig:cm_all}}
    \subfloat[Gradient norm (AIT).]{\includegraphics[align=c,width=.3\linewidth]{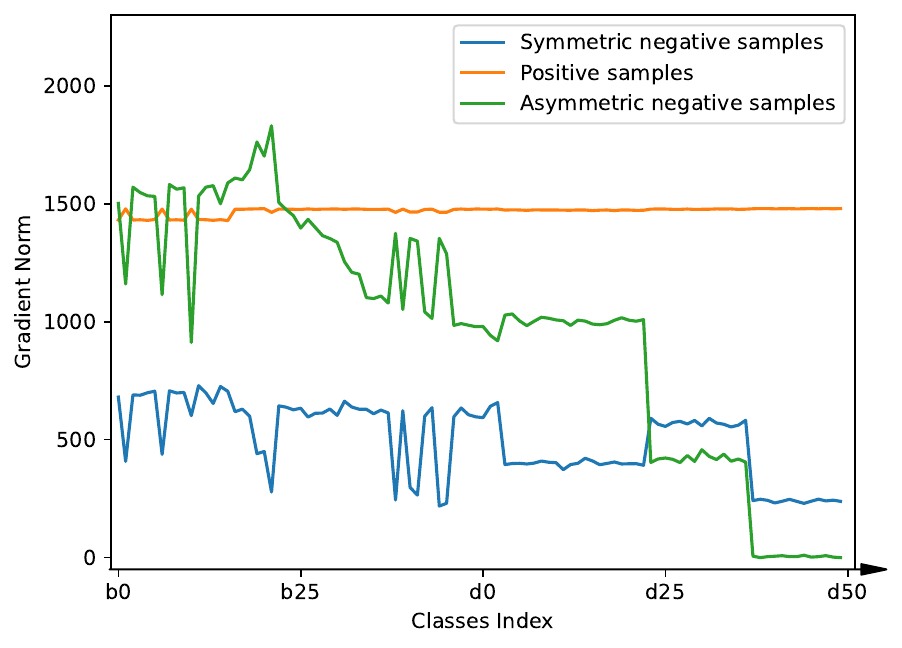}\label{subfig:gc_all}}
    \subfloat[Gradient norm timeline (AIT).]{\includegraphics[align=c,width=.3\linewidth]{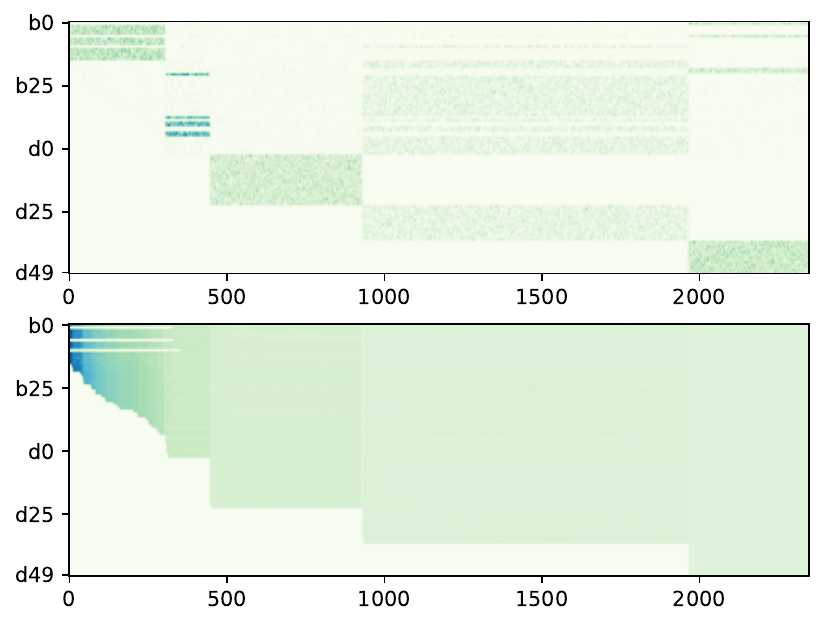}\label{subfig:gct_all}}
    \\
    \subfloat[Confusion matrix (SIT).]{\includegraphics[align=c,width=.3\linewidth]{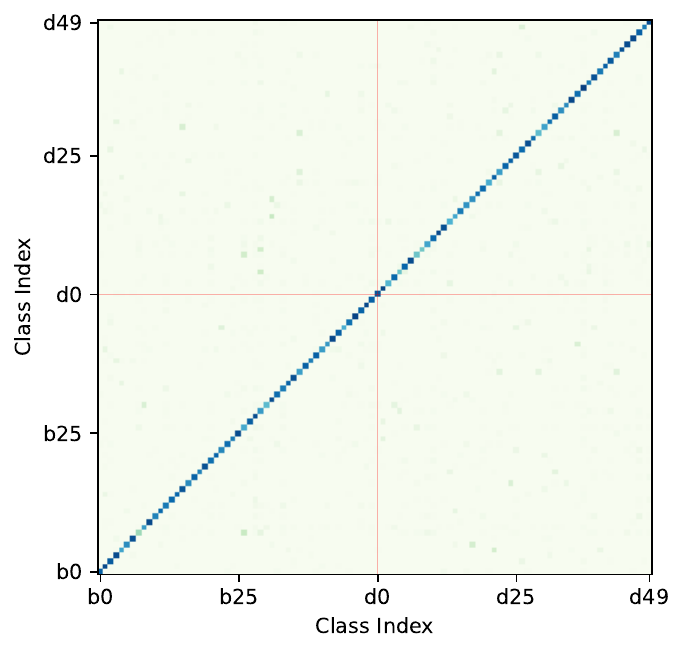}\label{subfig:cm_batch}}
    \subfloat[Gradient norm (SIT).]{\includegraphics[align=c,width=.3\linewidth]{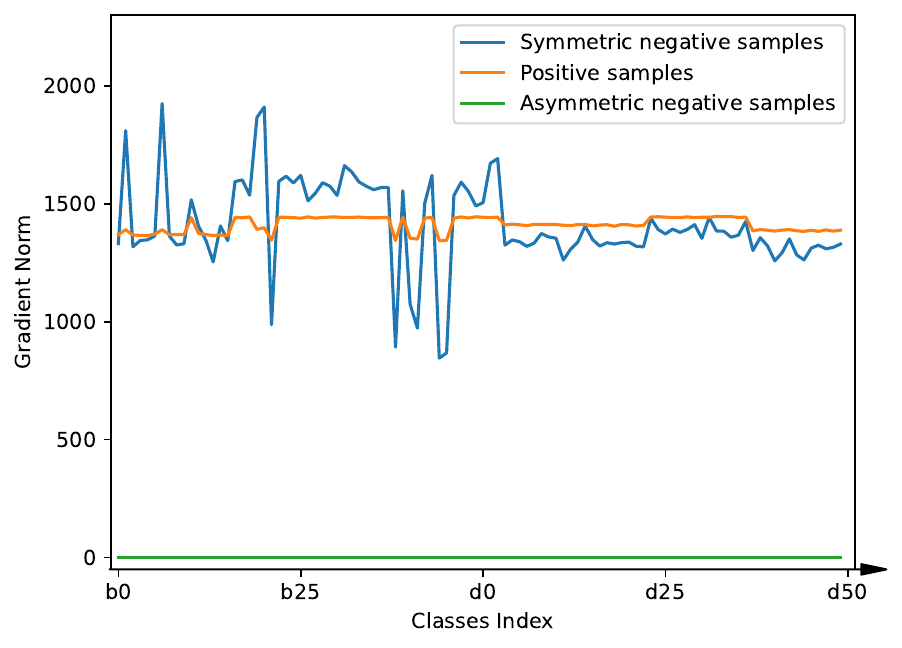}\label{subfig:gc_batch}}
    \subfloat[Gradient norm timeline (SIT).]{\includegraphics[align=c,width=.3\linewidth]{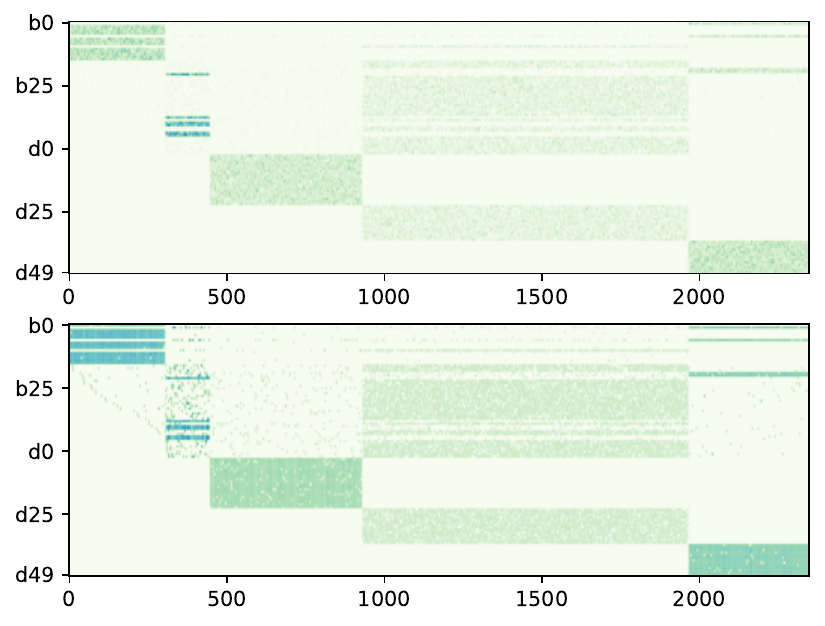}\label{subfig:gct_batch}}
    \caption{Analysis of Symmetric Image-Text tuning strategy. 
    % SIT stands for Symmetric Image-Text tuning strategy, and AIT represents Asymmetry Image-Text tuning strategy. 
    In the coordinate axis, \textit{b} denotes blurry classes, and \textit{d} represents disjoint classes. Furthermore, the classes are arranged according to their occurrence time. For the gradient timeline, the subplot at the top represents the gradient of the positive samples, while the subplot at the bottom represents the gradient of the negative samples. }
\label{fig:grad}
\end{figure*}

As discussed in Section \ref{sect:sit}, the asymmetry between image and text can lead to a degradation in the performance of OLL, a phenomenon we investigated through comparative experiments on CIFAR-100, TinyImageNet, and ImageNet-R, with results summarized in Table \ref{tab:sit}. In our experiments, the Asymmetric Image-Text training strategy (AIT) involved matching all previously seen classes with images at each time step, which, when compared to its baseline, led to a significant reduction in $\mathcal{A}_{\text{last}}$ while the $\mathcal{A}_{\text{auc}}$ is slightly lower than the baseline. Notably, as depicted in Figures \ref{subfig:cm_all} and \ref{subfig:cm_batch}, AIT resulted in the predictions bias towards new classes, indicative of catastrophic forgetting when contrasted with the SIT. To explore the reasons behind this phenomenon, we conducted an analysis of the gradients during training. Here, we considered the text encoder and text features collectively as a classifier and tallied the gradients of positive and negative samples. For Figures \ref{subfig:gc_all} and \ref{subfig:gc_batch}, we designated negative samples from the classes present at the current batch as symmetric negative samples, while those from classes not present are termed asymmetric negative samples. It is observed that for SIT, the gradients for both positive and negative samples are largely maintained at the same order of magnitude. In contrast, for AIT, the total gradient for negative samples across the OLL process is greater than that for positive samples for blurry classes, whereas for disjoint classes, the gradient for negative samples is less than that for positive samples. Moreover, the gradient for negative samples corresponding to new classes progressively diminishes. Figures \ref{subfig:gct_all} and \ref{subfig:gct_batch} illustrate the temporal evolution of the gradients. For AIT, the gradients on negative samples tend to become uniform, which hinder the CLIP model to effectively distinguish between new and old classes. The gradient analysis aligns with the observations in Figure \ref{subfig:cm_all}, where the model is inclined to predict classes as new. Overall, compared to AIT, SIT demonstrates the ability to selectively update knowledge, effectively mitigating catastrophic forgetting and preserving the generalization of CLIP model.

\begin{table}
\centering
\caption{Comparison the impact of training strategies on the performance of online lifelong learning. AIT represents the asymmetric image-text training strategy. }
\label{tab:sit}
\scalebox{0.85}{
\begin{tblr}{
  width = \linewidth,
  colspec = {Q[270]Q[125]Q[125]Q[125]Q[125]Q[125]Q[125]},
  cells = {c},
  row{1} = {c},
  row{2} = {c},
  column{1} = {r},
  cell{1}{1} = {r=2}{},
  cell{1}{2} = {c=2}{0.268\linewidth},
  cell{1}{4} = {c=2}{0.262\linewidth},
  cell{1}{6} = {c=2}{0.25\linewidth},
  hline{1,10} = {-}{0.1em},
  hline{3} = {-}{},
  hline{4} = {-}{dashed},
  hline{2} = {2-3}{leftpos = -1, rightpos = -1, endpos},
  hline{2} = {4-5}{leftpos = -1, rightpos = -1, endpos},
  hline{2} = {6-7}{leftpos = -1, rightpos = -1, endpos},
}
\textbf{ Method } & CIFAR-100 &  & TinyImageNet &  & ImageNet-R & \\
 & $\mathcal{A}_{\text{auc}}$ & $\mathcal{A}_{\text{last}}$ & $\mathcal{A}_{\text{auc}}$ & $\mathcal{A}_{\text{last}}$ & $\mathcal{A}_{\text{auc}}$ & $\mathcal{A}_{\text{last}}$\\
Continual-CLIP & 69.57±1.05 & 66.26±0.02 & 71.55±0.97 & 65.18±0.03 & 76.63±0.52 & 71.12±0.34\\
SIT-MaPLe & 69.89±2.69 & 67.61±1.01 & 71.17±1.45 & 65.78±1.90 & 78.50±0.74 & 73.48±1.00\\
AIT-MaPLe & 65.77±4.07 & 47.12±8.46 & 67.89±2.00 & 51.00±13.64 & 78.17±0.90 & 78.98±3.64\\
SIT-Adapter & 81.03±2.00 & 79.15±2.08 & 80.30±1.19 & 74.96±1.44 & 83.28±1.01 & 78.76±0.39\\
AIT-Adapter & 68.80±3.68 & 53.36±5.37 & 69.58±1.23 & 51.92±5.56 & 78.36±0.37 & 66.36±2.57\\
SIT-LoRA & 81.77±1.80 & 80.80±1.26 & 80.64±1.18 & 75.08±1.09 & 84.15±0.38 & 79.35±0.35\\
AIT-LoRA & 72.31±5.18 & 53.61±4.82 & 69.40±1.39 & 50.12±4.73 & 79.01±0.60 & 67.31±5.90\\
\end{tblr}}
\end{table}

% 4.text encoder与image encoder的比较以及几种PET方法间两种训练范围的比较
\subsubsection{Analysis of the PET methods on CLIP model.}

\begin{table}
\centering
\caption{Comparison the effects of PET on the performance of online lifelong learning and the generalization of the CLIP model. {\#P} represents the trainable parameters. }
\label{tab:encoder}
\scalebox{0.65}{
\begin{tblr}{
  cells = {c},
  column{1} = {r},
  cell{1}{1} = {r=2}{},
  cell{1}{2} = {r=2}{},
  cell{1}{3} = {c=2}{},
  cell{1}{5} = {c=6}{},
  hline{2} = {2-9}{},
  hline{1,11} = {-}{0.1em},
  hline{3} = {-}{},
  hline{4} = {-}{dashed},
  hline{2} = {3-4}{leftpos = -1, rightpos = -1, endpos},
  hline{2} = {5-10}{leftpos = -1, rightpos = -1, endpos},
}
\textbf{ Method } & {\#P} & \textbf{TinyImageNet} &              & \textbf{Target} &             &              &             &              &             \\
                  & & $\mathcal{A}_{\text{auc}}$                   & $\mathcal{A}_{\text{last}}$         & Food101         & Caltech101  & EuroSAT      & Flowers102  & OxfordPets   & Average     \\
Continual-CLIP &   0.00 M       & 71.55±0.97    & 65.18±0.03     & 90.66        & 95.40           & 60.03        & 74.83
           & 94.12           & 83.01         \\
SIT-MaPLe &   1.19 M      & 71.06±0.38              & 65.61±2.83               & 85.65±0.92        & 90.53±3.51           & 45.92±4.28        & 60.65±8.12           & 79.32±4.73                   & 72.41±4.19                \\
SIT-Adapter &   1.98 M          & 80.41±1.07    & 75.58±1.11     & 76.11±3.45        & 92.10±0.33           & 22.84±2.74        & 54.61±2.41           & 79.48±1.08           & 65.03±1.53         \\
SIT-Adapter (image only) & 1.19 M & 79.26±0.81    & 73.05±0.90     & 73.62±3.94        & 91.13±1.13           & 22.51±1.88        & 53.38±7.50           & 82.17±1.74           & 64.56±2.63         \\
SIT-Adapter (text only)& 0.79 M & 76.15±0.91    & 70.59±0.93     & 80.74±3.05        & 91.04±1.03           & 45.27±5.13        & 61.55±2.59           & 79.27±5.51           & 71.57±1.50         \\
SIT-LoRA &     0.37 M           & 80.82±0.79    & 75.85±1.19     & 79.76±1.20        & 92.71±1.48           & 24.85±3.01        & 59.62±1.82           & 80.65±3.61           & 67.52±0.75         \\
SIT-LoRA (image only) &  0.22 M   & 79.53±0.73    & 73.33±1.09     & 79.46±1.11        & 91.01±0.93           & 24.92±1.77        & 60.06±1.82           & 82.42±3.45           & 67.57±0.83         \\
SIT-LoRA (text only) & 0.15 M   & 76.23±0.96    & 70.61±0.59     & 79.43±2.03        & 90.69±1.12           & 42.18±5.89        & 57.24±3.88           & 77.53±5.38           & 69.41±2.39        
\end{tblr}}
\end{table}

% 梯度分析+混淆矩阵可视化（tinyimagenet，task=10，N=50，M=10）

% \begin{table}
% \centering
% \caption{Comparing the computational cost of different PET methods in terms of training parameters and GPU burden during training.}
% \label{tab:gpu}
% \begin{tblr}{
%   cells = {c},
%   column{1} = {r},
%   hline{1,6} = {-}{0.1em},
%   hline{2} = {-}{},
% }
% \textbf{ Method } & \textbf{ GPU } & \textbf{ Total Params } & \textbf{ Trainable Params } \\
% Continual-CLIP    & N/A            & 149.6 M                 & 0 M                         \\
% SIT-MaPLe         & 6048 MiB       & 150.8 M                 & 1.1 M                       \\
% SIT-Adapter       & 10728 MiB      & 151.6 M                 & 2.0 M                       \\
% SIT-LoRA          & 12380 MiB      & 150.0 M                 & 0.4 M                       
% \end{tblr}
% \end{table}

In this experiments, we conducted a comparative analysis of various PET methods to evaluate their impact on the performance and generalizability in OLL. As evident from Tables \ref{tab:si-blurry} and \ref{tab:encoder}, MaPLe, which is based on prompt tuning, shows minimal improvement over the baseline in OLL but also has the least detrimental effect on generalization compared to other methods. Both LoRA and Adapter demonstrate significant performance in OLL. LoRA has a relatively smaller impact on generalization, which has fewer parameters. Furthermore, we compared the effects of only tuning the text encoder versus the image encoder. Table \ref{tab:encoder} indicates that tuning the image encoder outperform only tuning the text encoder in OLL, implying that the model needs to adapt to different image distributions throughout lifelong learning. In contrast, for zero-shot learning, only tuning the text encoder is more effective than the image encoder, suggesting that zero-shot learning benefits more from the rich semantic information contained in language. These findings underscore the importance of selecting the appropriate tuning strategy based on the learning paradigm. In OLL, where the model encounters a continuous stream of new data, it is crucial to focus on the image encoder to adapt to varying image distributions. Conversely, in zero-shot learning, leveraging the capacity of text encoder to extract semantic information is more beneficial.

\begin{table}
\centering
\caption{Comparison the impact of batch size in online lifelong learning.}
\label{tab:batchsize}
\scalebox{0.7}{
\begin{tblr}{
  cells = {c},
  column{1} = {r},
  cell{1}{1} = {r=2}{},
  cell{1}{2} = {c=2}{},
  cell{1}{4} = {c=6}{},
  hline{3} = {-}{},
  hline{2} = {2-9}{},
  hline{1,9} = {-}{0.1em},
  hline{3} = {-}{},
  hline{4} = {-}{dashed},
  hline{2} = {2-3}{leftpos = -1, rightpos = -1, endpos},
  hline{2} = {4-9}{leftpos = -1, rightpos = -1, endpos},
}
\textbf{ Method } & \textbf{TinyImageNet} &              & \textbf{Target} &             &              &             &              &             \\
                  & $\mathcal{A}_{\text{auc}}$                   & $\mathcal{A}_{\text{last}}$         & Food101         & Caltech101  & EuroSAT      & Flowers102  & OxfordPets   & Average     \\
Continual-CLIP    & 71.55±0.97    & 65.18±0.03     & 90.66        & 95.40           & 60.03        & 74.83
           & 94.12           & 83.01         \\
8                     & 74.44±0.90    & 63.85±0.87     & 67.55±3.34        & 89.34±0.88           & 18.47±3.26        & 41.43±6.56           & 67.17±4.71           & 56.79±2.07         \\
16                    & 78.29±0.29    & 70.06±1.33     & 76.19±1.36        & 90.87±0.22           & 17.92±1.41        & 53.96±1.50           & 75.64±0.64           & 62.92±0.70         \\
32                    & 80.20±0.50    & 73.58±0.31     & 80.24±1.08        & 92.07±0.42           & 25.93±4.33        & 58.18±1.29           & 77.99±2.17           & 66.88±1.52         \\
64                    & 80.91±0.46    & 74.91±0.30     & 81.62±0.58        & 92.57±0.42           & 24.34±4.09        & 58.97±3.12           & 81.70±3.44           & 67.84±0.50         \\
128                   & 80.83±0.52    & 75.50±0.22     & 82.89±0.56        & 92.98±0.15           & 23.79±3.35        & 60.43±0.91           & 80.79±0.36           & 68.18±0.76        
\end{tblr}}
\end{table}

\subsubsection{Analysis of the impact of batch size in online lifelong learning.}

The objective of this experiment is to determine how variations in batch size affects the efficiency and effectiveness of the learning process in an OLL scenario. We conducted a series of experiments with batch sizes ranging from 8 to 128 and use the SIT-LoRA within the Si-Blurry setting on the tinyImagenet dataset. The results, as indicated in Table \ref{tab:batchsize}, demonstrate that the overall impact of batch size on OLL performance is relatively minor, with the $\mathcal{A}_{\text{auc}}$ differing by only 6.39\% between the smallest and largest batch sizes tested. Notably, the final performance in OLL converges when the batch size exceeds 16, suggesting that beyond this point, increasing the batch size does not yield significant improvement in performance. Interestingly, for zero-shot testing, an increase in batch size during OLL is found to mitigate the reduction in the generalizability of CLIP model, indicating that larger batches may contribute to better retention of knowledge about novel classes. These findings suggest that while batch size does play a role in OLL, the optimal size for maintaining a balance between performance and generalizability may not need to be exceedingly large.

\section{Conclusion}

Online lifelong learning entails the capability of models to learn from data streams without preset constraints such as the total number of classes or maximum memory capacity, and to apply and evaluate model performance at any given moment. In this paper, we proposed that vision-language models such as CLIP are more suitable online lifelong learners compared to traditional image classification models. Through gradient analysis, we discovered that asymmetry between text and image during PET in OLL can lead to catastrophic forgetting. To address this, we introduced a simple yet effective strategy known as the Symmetric Image-Text tuning strategy, which matches images and class labels within the current batch only during online learning. We conducted extensive experiments in both OLL and CIL scenarios and evaluated the impact of lifelong learning on generalizability of CLIP. Our future work will explore Mixture of Experts (MoE) to accumulate knowledge gained throughout lifelong learning while avoiding catastrophic forgetting of vision-language models.

\bibliographystyle{unsrtnat}
\bibliography{main}

%%%%%%%%%%%%%%%%%%%%%%%%%%%%%%%%%%%%%%%%%%%%%%%%%%%%%%%%%%%%

% \appendix

% \section{Appendix / supplemental material}

% Optionally include supplemental material (complete proofs, additional experiments and plots) in appendix.
% All such materials \textbf{SHOULD be included in the main submission.}

%%%%%%%%%%%%%%%%%%%%%%%%%%%%%%%%%%%%%%%%%%%%%%%%%%%%%%%%%%%%

% \input{checklist}

\end{document}